# Deep learning-based approach for tomato classification in complex scenes

Mikael A. Mousse[1,*], Bethel C. A. R. K. Atohoun[2], and Cina Motamed[3]

[1]Institut Universitaire de Technologie, Université de Parakou, Parakou, Bénin
[2]Ecole Supérieure de Gestion d'Informatique et des Sciences, Cotonou, Bénin
[3]Université d'Orléans, Orléans, France
Email: mikael.mousse @univ-parakou.bj

*Abstract*—**Tracking ripening tomatoes is time consuming and labor intensive. Artificial intelligence technologies combined with those of computer vision can help users optimize the process of monitoring the ripening status of plants. To this end, we have proposed a tomato ripening monitoring approach based on deep learning in complex scenes. The objective is to detect mature tomatoes and harvest them in a timely manner. The proposed approach is declined in two parts. Firstly, the images of the scene are transmitted to the pre-processing layer. This process allows the detection of areas of interest (area of the image containing tomatoes). Then, these images are used as input to the maturity detection layer. This layer, based on a deep neural network learning algorithm, classifies the tomato thumbnails provided to it in one of the following five categories: green, brittle, pink, pale red, mature red. The experiments are based on images collected from the internet gathered through searches using tomato state across diverse languages including English, German, French, and Spanish. The experimental results of the maturity detection layer on a dataset composed of images of tomatoes taken under the extreme conditions, gave a good classification rate.**

## I. INTRODUCTION

One of the most important aspects for consumers in the agricultural industry is product quality. Traditionally, the inspection process is done manually, which is time-consuming, subjective and unreliable. For this reason, a great effort is made by the scientific community to develop automatic systems that help to improve the inspection process, from time consumption and consistency points of view. Since quality control influences the viability of products, countries producing agricultural raw materials have invested significant research efforts in the automated monitoring and control of crop growth [1]. Tomato maturity classification is crucial for optimizing harvesting and post-harvest processes in the agricultural industry.

Tomato classification, a key task in agricultural automation, has witnessed remarkable advancements with the integration of machine learning techniques. In recent years, researchers have employed various methodologies to enhance the accuracy and efficiency of tomato classification systems. All approaches can be classified into three categories : Traditional approaches, advanced approaches and hybrid approaches. The traditional approaches regroup visual inspection-based methods and physical properties-based methods. The advanced approaches are based on image processing, computer vision spectroscopy and hyperspectral imaging processing. The hybrid approaches combine at least two of the previous approaches.

The next section of this paper presents the related works.

The third section shows the proposed approach. The fourth section presents the experiment setup. In this section, we describe the dataset and the experimentation environment. The fifth section discusses the performance of the proposed system. Finally, we end the paper with a conclusion.

## II. RELATED WORKS

This work is focused on the third category of algorithm. Zhu et al. [2] presented an automated multi-class classification approach for measuring and evaluating the maturity of tomatoes through the study and classification of the different stages of maturity of both optical parameters and their combinations to classify the tomatoes into different ripeness grades. The specific objectives of their work are to :

- Measure the optical absorption and scattering coefficients of 'Sun Bright' tomatoes with different ripeness grades over 500–950 nm, using a hyperspectral imaging-based spatially-resolved instrument, and evaluate their relationship with the ripeness of tomatoes;
- Develop discriminant models for classification of tomatoes into either six or three ripeness grades, using the absorption and scattering coefficients and their combinations.

Luna et al. developed a convolutional neural network-based solution that allows disease detection in tomato plants. A motorized image capture box was designed to capture all four sides of each tomato plant to detect and recognize leaf diseases. A specific breed of tomato which is Diamante Max was used as the test subject. The system was designed to identify diseases such as Phoma rot, leaf miner and target spot. Using a dataset of 4,923 leaf images from diseased and healthy tomato plants collected under controlled conditions, they trained a deep convolutional neural network to identify three diseases or lack thereof [3]. Hu et al. [4], proposed a method for detecting ripe tomatoes using a vision system. The purpose of this method is to evaluate the feasibility of combining deep learning with edge segmentation to detect individual tomatoes in complex environments with a view to avoiding confusions of overlapping tomatoes. Castro et al. combined four supervised machine learning algorithms namely ANN, DT, SVM and KNN as well as three color spaces (RGB, HSV and L*a*b*) for Cape gooseberry classification according to their level of maturity. The objective was to find the best combination of supervised learning technique and color space for classification. To do this, they collected 925 Cape gooseberry fruit samples and categorized them into seven classes according to the level of ripeness after extracting information about the color parameters of the three-color spaces from each fruit sample [5].

To overcome the difficulties of detection by artificial vision in such cases, some studies such as the work [6] proposed a detection method based on color and shape characteristics. Haggag et al. applied supervised and unsupervised neural learning and deep learning algorithms to three different sets of tomato images with hundreds of iterations to identify the best techniques and network configurations. more efficient. These are the convolutional neural network, the artificial neural network, the self-organizing map (SOM), the learning vector quantization and the support vector machine [7]. To automatically identify the maturity of tomatoes, Huang et al. [8] proposed a Mask R-CNN fuzzy model. The aim of their study is to enable farmers to avoid losses due to late and early harvests. They used the fuzzy c-means model for identification and segmentation of acquired tomato images to maintain image foreground and background spatial information. In their work, Xie et al. [9] constructed a dataset of 20 categories of fruits and vegetables in 11 different states ranging from solid, sliced to creamy paste. This dataset was constructed with 11943 images uploaded using the Google search engine, manually reviewed and then categorized based on food identity and then condition. The convolutional neural network architectures allowed them to perform the category, state and category recognition and food state tasks. To overcome the lack of labeled data, they exploited artisanal features as well as deep features extracted from CNNs combined with support vector machines as an alternative to end-to-end classification. Ni et al. [10] proposed a system for monitoring the process of changing banana freshness using transfer learning. To build the dataset, they selected 103 bananas from two varieties. In order to meet the data set size requirements of transfer learning, they amplified the data following the data amplification techniques such as rotation, translation and mirroring to increase the number of datasets. The dataset was then split into a training and validation set. Most of these algorithms failed in the case of complex scenes. Indeed actually, there are two principal challenges with automatic fruit detection or recognition. One is that complex scenes. In these scenes we need to address the problems such as backlighting, direct sunlight, overlapping fruit and branches, blocking leaves, etc. All these phenomena represent obstacles to the detection and recognition of fruits.

To avoid this problem, Faisal et al. [11] proposed a decision system using computer vision and deep learning techniques to detect seven different maturity stages/levels of date fruit. These stages are Immature stage 1, Immature stage 2, Pre-Khalal, Khalal, Khalal with Rutab, Pre-Tamar, and Tamar. Working in this direction too and focusing on automatic tomato recognition, Xu et al. [12] proposed a fast method of detecting tomatoes in a complex scene for picking robots. They used an image enhancement algorithm to improve the detection ability of the algorithm in complex scenes. Finally, they designed several groups of comparative experiments to prove the rationality and feasibility of this method. Das et al. [23] introduced machine learning based algorithm for tomato maturity grading. The performance of their system is assessed on the real tomato datasets collected from the open fields using Nikon D3500 CCD camera. Filoteo-Razo et al. presented a non-invasive and low-cost optoelectronic system for detecting color changes in oranges to predict the ripening stage [24]. Deepkiran et al. [25] developed detection system for plant-disease using convolution neural network (CNN) model. A more comprehensive review on detection and classification of plant diseases using machine learning is presented in [26]. Some research works [13]-[16] suggest to generate candidate regions. These regions are classified to detect the fruit.

The main contribution of this work is to propose a strategy for tomato state recognition in the complex scenes. By comparing our work to state-of-the-art algorithms, we note that after the tomato detection phase, we classify the state of maturity of the tomato. The first part of our work consists in proposing a new competitive strategy for tomato detection. The proposed approach uses a preprocessing layer to segment the image into regions. The objective of this segmentation is to extract the thumbnails containing the parts of interest for the study. Once the tomatoes are detected, we use an algorithm based on deep neural networks for classification. Five classes have been identified for this work.

## III. PROPOSED METHOD

The system that we propose is composed of two main layers. The first layer is image segmentation. The objective of this layer is to isolate the parts of the image which contain tomatoes. This first step is very important because if the contours of the tomato are not well realized, the final system will not be efficient. After the segmentation of the tomato image, the result obtained is inserted into the maturity status recognition system. This system, based on deep learning, is responsible for classifying tomatoes in one of the classes of our study.

### A. Image segmentation

Image segmentation is also important for tomato state recognition as it allows for the precise delineation and extraction of various tomato regions or components from an image. By segmenting the tomato, different aspects such as color, shape, texture, and defects can be analyzed, leading to effective state recognition. The flowchart of the proposed tomato segmentation is illustrated by Figure 1.

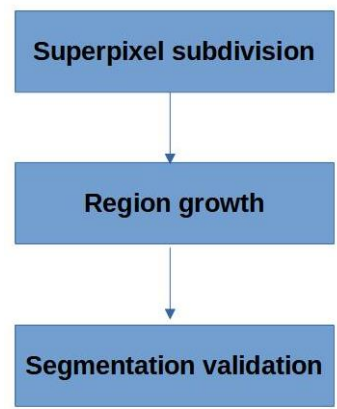

Figure 1. Flowchart of proposed tomato image segmentation

According to Figure 1, we have three important steps. The objective of the first step is to divide the image into compact and perceptually homogeneous superpixels. The second step allows us to iteratively grow the region by incorporating neighboring superpixels based on predefined similarity criteria. Finally, we evaluate each detected region in order to find the regions which contain tomatoes.

### 1) Superpixel subdivision

The use of superpixels is topical in the realization of computer vision-based systems. Superpixel segmentation is the simplification of the image into a number of homogeneous regions. Each region is a superpixel and the pixels of each region have very similar characteristics. The number of superpixels is quite large but significantly lower than the number of pixels. Of our many superpixel grouping strategies exist. But we chose the one proposed by Schick et al. [17]. The performance of this superpixel segmentation is shown on some computer vision-based applications [18]. The pseudo code of their strategy is given by algorithm 1.

| Algorithm 1 : superpixels segmentation |
|---|
| 1 Initialize cluster centers   by  sampling pixels at regular grid steps. |
| 2 Perturb cluster centers in a neighborhood, to the lowest gradient position using expression. |
| 3 **repeat** |
| 4      **for** each cluster center  do |
| 5         Assign the best matching pixels from an n*n neighborhood around cluster center according to the distance measure. |
| 6      Compute new cluster centers and residual error {distance between previous centers and recomputed centers}. |
| 7 **until**  <= threshold |
| 8 Enforce connectivity. |

### 2) _Region growth

Region growth is an image segmentation technique used to partition an image into meaningful regions or regions of interest  based on certain similarity criteria. The basic idea behind region growth is to start with seed points or regions and iteratively  grow  these  regions  by  incorporating neighboring pixels or regions that have similar characteristics. In this work, we use the CSP regional growth method proposed by Tao et al. [19]. This approach was inspired by the metabolic ecology theory. Experimental results have shown that this approach has good accuracy in forests. The CSP regional growth method is declined in three important parts: points normalization, trunk detection and diameter at breast height estimation, and finally segmentation.

### 3) Segmentation validation

The segmentation validation is important in order to identify the parts which contain a tomato. In this work, our strategy for segmentation validation is inspired by the work of Xu et al. [12]. This strategy is presented by Figure 2.

As shown in Figure 2, the tomato recognition process is performed in two steps. The first step is model training. This step requires the tomato dataset and the boundary box labels to be fed into the convolutional neural network; iterative training is then conducted, and a fully trained model is obtained. The second step is model inference. At this stage, the potential regions detected after the region growth process are input into the trained model. The goal is to find if these regions are tomatoes or not.

### B. Tomato state classification

In our research endeavor, we have constructed a classification framework by harnessing the capabilities of recurrent neural networks (RNN) integrated with an attention mechanism. We have meticulously compared the performance of this approach with traditional and Bayesian-type methodologies. Our innovative system comprises three interconnected modules. These modules encompass the preprocessing of inputs, enabling the extraction of vital features, the establishment of feature embeddings to optimize the preparation of model inputs, and, lastly, the classifier itself. Our classifier is composed of a flexible and trainable features layer meticulously crafted with LSTM cells. In addition, we have diligently employed the layers of a stacked single head attention recurrent neural network, seamlessly integrating the Boom layer and culminating in a softmax classifier.

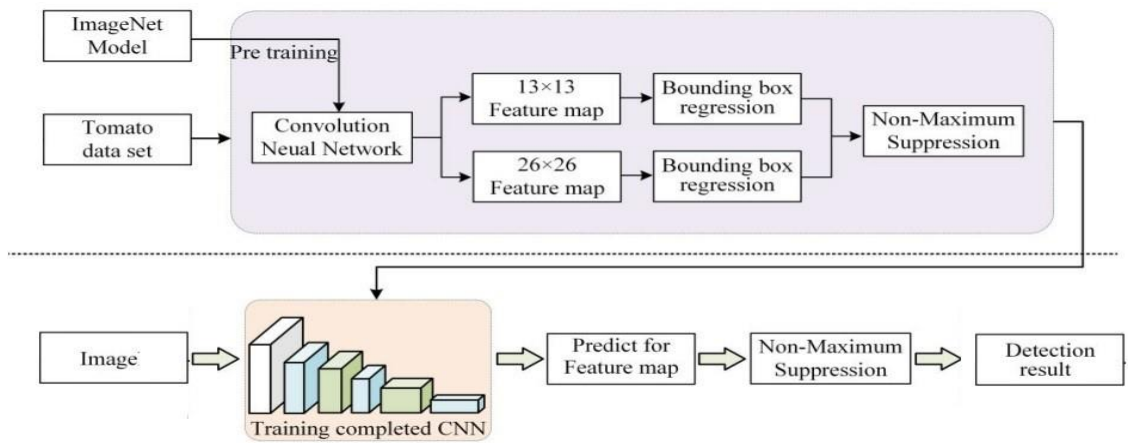

Figure 2. The training and detecting process of the tomato detection method.

Diverging from the established norms of transformer-based approaches as documented in previous literature [20], our final model gracefully adapts a distinctive strategy with the use of a single attention head. This novel simplification, inspired by the pioneering work of [21], has convincingly demonstrated superior performance in terms of effectively capturing and harnessing crucial informational aspects. Consequently, we have judiciously opted against cluttering our model with a multitude of attention heads, as their true efficacy remains uncertain within the context of the less intricate signals that we meticulously manipulate throughout the course of this research. It is essential to underscore that our attention mechanism showcases exemplary computational efficiency, largely due to the judicious application of matrix multiplication, which exclusively operates on the query as elegantly described in [21].

The incorporation of the boom layer, which has proved to be remarkably advantageous within our framework, offers an effective means to regulate the exponential expansion of vector dimensions emanating from the attention layers. To achieve this, we employ a projection strategy that elevates the vectors to a higher dimensionality before gracefully reverting them back to their original dimensions. To best capture the intrinsic characteristics of tomato image sounds, our model is thoughtfully imbued with large feed-forward layer steps at various strategic points, allowing us to dynamically encode the multifaceted features across each distinct feature dimension. This astute encoding process results in the generation of dynamic vectors embedded within hidden states, further enhancing the model's capacity to extract meaningful representations. The model is presented by Figure 3.

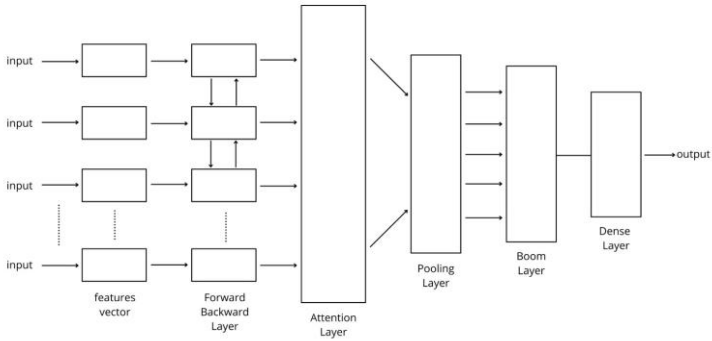

Figure 3. Classification model architecture.

## IV. RESULTS AND DISCUSSION

Suitable datasets are indispensable throughout all phases of computer vision-based research, spanning from the initial training phase to the subsequent evaluation of recognition algorithm effectiveness. For this work, the images compiled for the dataset were sourced from the Internet, gathered through searches using tomato state across diverse languages including English, German, French, and Spanish. These images, categorized into five distinct classes. These images are taken in different conditions (side light, back light, sunlight, day, evening, night).

After the collection of the image, the dataset was augmented with additional images. The primary objective of this step is to obtain consequent images in order to train the network in discerning distinctive features that differentiate between various classes. By incorporating a larger set of augmented images, the network's ability to grasp pertinent features has been amplified. As a culmination, a database comprising 50910 images for training and 3000 images for validation was established. The additional images are obtained using the strategy proposed by Stearns and Kannapan [22]. The experiments are performed using a machine with an Intel Core™i7 CPU and NVIDIA GTX 1050ti GPU. The operating system is Windows 10. This machine is used to build a TensorFlow deep learning framework. The training and detection of the tomato object detection network model was programmed in Python.

To prove the efficacy of the proposed approach, we use some metrics. These metrics are precision, f1-score and execution time. The calculation of precision (P) and f1-score (F1) are presented in formula (1).

Suitable datasets are indispensable throughout all phases of computer vision-based research, spanning from the initial training phase to the subsequent evaluation of recognition algorithm effectiveness. For this work, the images compiled for the dataset were sourced from the Internet, gathered through searches using tomato state across diverse languages including English, German, French, and Spanish. These images, categorized into five distinct classes. These images are taken in different conditions (side light, back light, sunlight, day, evening, night).

After the collection of the image, the dataset was augmented with additional images. The primary objective of this step is to obtain consequent images in order to train the network in discerning distinctive features that differentiate between various classes. By incorporating a larger set of augmented images, the network's ability to grasp pertinent features has been amplified. As a culmination, a database comprising 50910 images for training and 3000 images for validation was established. The additional images are obtained using the strategy proposed by Stearns and Kannapan [22]. The experiments are performed using a machine with an Intel Core™i7 CPU and NVIDIA GTX 1050ti GPU. The operating system is Windows 10. This machine is used to build a TensorFlow deep learning framework. The training and detection of the tomato object detection network model was programmed in Python. Figure 4 presents three images of the dataset and the tomato classification results.

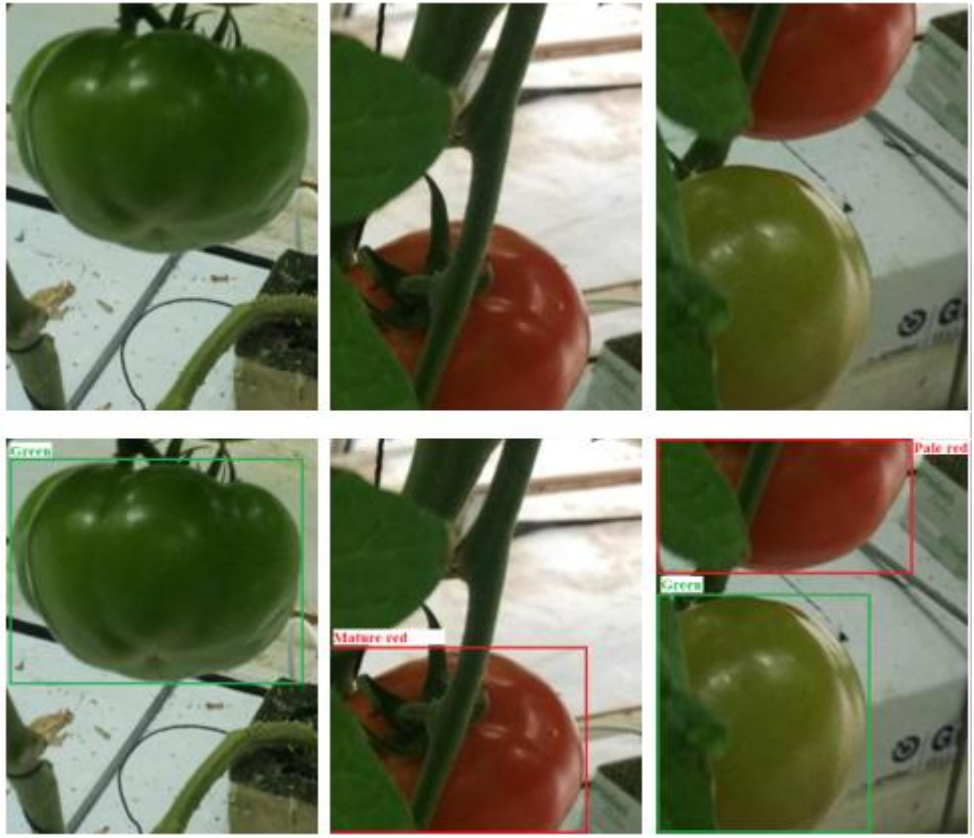

Figure 4. Tomato classification results.

To prove the efficacy of the proposed approach, we use some metrics. These metrics are precision, f1-score and execution time. The calculation of precision (P) and f1-score (F1) are presented in formula (1).

$$P = TP / (TP + FP)$$
$$R = TP / (TP + FN) \quad (1)$$
$$F1 = 2*P*R / (P + R)$$

In formula (1), P is the precision rate, R is the recall rate, TP is the number of true positive samples, FP is the number of false positive samples, and FN is the number of false negative samples. Using these metrics, we compare our proposed method to other algorithms of state of the art These algorithms are Xu et al. [12], Ren et al. [13], He et al. [14] and Dai et al. [15]. The results of comparison are consigned in Table 1.

Table 1. Comparison of test results.

| Method | Precision | F1 score |
|--------|-----------|----------|
| Xu et al. [12] | 91.60 | 92.32 |
| Ren et al. [13] | **94.15** | **96.83** |
| He et al. [14] | 90.22 | 91.33 |
| Dai et al. [15] | 88.16 | 90.15 |
| Our Method | 93.38 | 95.77 |

According to Table 1, we show that using the proposed approach we achieve an acceptable precision for the tomato detection process. The F1 score also proves the efficiency of the proposed approach.

After this step, we can test the efficiency of the classification algorithm. Upon refining the network's parameters, a comprehensive accuracy level of 97.06% was attained by the 100th training iteration (as opposed to 94.9% prior to fine-tuning). Remarkably high accuracy results were already obtained, accompanied by notably diminished loss, even as early as the 30th training iteration. However, a compelling equilibrium between accuracy and loss was consistently maintained for high accuracy outcomes beyond the 60th iteration. Taking into account the distribution by class, we obtain Table 2.

Table 2. Result of tomato state classification

| Classes | | |
|---------|---|---|
| Class1 | All images | 800 |
| | **Accurately predicted** | **787** |
| Class2 | All images | 760 |
| | **Accurately predicted** | **735** |
| Class3 | All images | 425 |
| | **Accurately predicted** | **413** |
| Class4 | All images | 390 |
| | **Accurately predicted** | **375** |
| Class5 | All images | 625 |
| | **Accurately predicted** | **602** |
| Total | All images | 3000 |
| | **Accurately predicted** | **2912** |

According to the table 2, we obtain a percentage of good classification pf each class. These percentages are reported in the Figure 5.

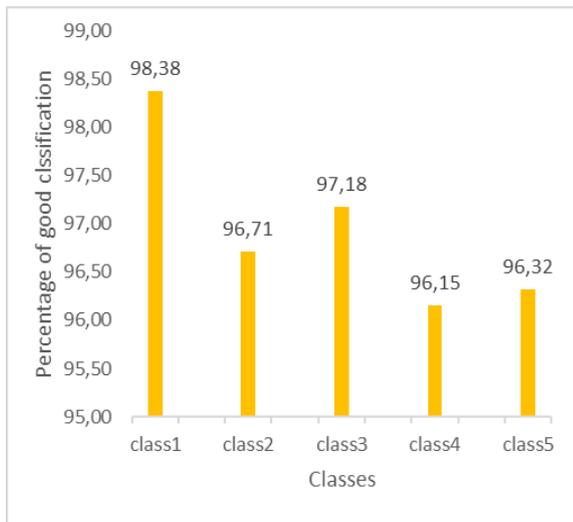

Figure 5: Percentage of good classification

Using Figure 5, we can observe that the lowest rate of good classification is 96.15% and this is obtained with images of class 4. These values have increased reaching up to 98.38% for class 1 images.
According to these values, we can conclude that our proposed strategy has competitive results for tomato classification.

## V. CONCLUSION

There are many automated or computer vision tomato detection and classification methods, but this area of research is still lacking. In this paper, we proposed a deep learning-based approach for tomato classification in a complex environment. The proposed approach takes place at two levels. The first level segments tomatoes from images presenting significant challenges (lack of light, changing weather, etc.). This segmentation is done using a pre-processing algorithm whose role is to remove noise and put homogeneous pixels together. The second level makes it possible to classify the state of the tomato according to its maturity. In this study we made a classification in five classes: green, brittle, pink, pale red, mature red. The results of the experiment showed the competitiveness of the proposed strategy.

Moreover, forthcoming endeavors will encompass expanding the model's application by training it to recognize the state of tomatoes across broader expanses of land. This will involve amalgamating aerial photographs taken by drones from orchards and vineyards with convolutional neural networks for precise object detection. Through the extension of this research, the authors aspire to make a meaningful contribution to sustainable development, influencing crop quality to benefit generations to come.

## CONFLICT OF INTEREST

The authors declare no conflict of interest.